%% file: nips_2017.tex
\documentclass{article}

\usepackage[final,nonatbib]{nips_2017}

\usepackage[utf8]{inputenc} %
\usepackage[T1]{fontenc}    %
\usepackage{hyperref}       %
\usepackage{url}            %
\usepackage{booktabs}       %
\usepackage{amsfonts}       %
\usepackage{nicefrac}       %
\usepackage{microtype}      %
\usepackage{epsfig}
\usepackage{graphicx}
\usepackage{caption}
\usepackage{subcaption}
\usepackage{float}
\usepackage{multirow, eucal}
\usepackage{color}

\def\P{{\mathbb P}}
\def\N{{\mathbb N}}
\newcommand{\etc}{\textit{etc.}}
\newcommand{\ie}{\textit{i.e.}}
\newcommand{\etal}{\textit{et al.}}

\title{Care about you: towards large-scale human-centric visual relationship detection}

\author{
  Bohan Zhuang, Qi Wu, Chunhua Shen\thanks{
    The first two authors contributed equally. C. Shen is the corresponding author.
  }, Ian Reid, Anton van den Hengel
\\
The University of Adelaide, SA 5005, Australia
}

\begin{document}

	\maketitle

	\begin{abstract}
		Visual relationship detection aims to capture interactions between pairs of objects in images.
Relationships between objects and humans represent a particularly important subset of this problem, with implications for challenges such as understanding human behaviour, and identifying affordances, amongst others.
	In addressing this problem we first construct a large-scale human-centric visual relationship detection dataset (HCVRD), which provides many more types of relationship annotation (nearly 10K categories) than the previous released datasets.
	This large label space better reflects the reality of human-object interactions, but gives rise to a long-tail distribution problem, which in turn demands a zero-shot approach to labels appearing only in the test set.  This is the first time this issue has been addressed.
	We propose a webly-supervised approach to these problems and demonstrate that the proposed model provides a strong baseline on our HCVRD dataset.
	\end{abstract}

	\input{intro.tex}
	\input{relwork.tex}
	\input{data.tex}

        \input{method.tex}
        \input{exp.tex}

\section{Conclusion}
We have
proposed a large-scale human-centric visual relationship detection dataset, which
is significantly larger and broader than previous datasets.  Human-centric relationships represent an important subclass of all relationships, not only because the human has agency, but also due to their practical importance for other challenges.  Increasing the scale of data available better captures the reality of the task, but rises to two important practical problems, the long-tail distribution issue and the zero-shot problem.
Motivated by the practical importance of the task, our methods address the issues raised in long-tail datasets and provide a strong baseline for further works based on our HVCRD dataset and similar data.

	\bibliographystyle{abbrv}
	{
		\small
		\bibliography{reference_2}
	}

\end{document}

%% file: intro.tex
\section{Introduction}

The challenge in visual relationship detection \cite{li2017vip, Liang2017VRD, lu2016visual} is to capture interactions between pairs of objects in an image. %
 In this paper, rather than detect interactions between arbitrary objects, we focus on capturing the relationships  between a human 
and an object. Recognising human-object relationships is a 
problem of significant practical import, and a subtly different challenge, than the more general case.  Humans have a far wider variety of modes of interaction than do general objects, but they also have agency, meaning that more can be drawn from human-object interactions than from other interactions.
For example, a human can interact with a bicycle in multiple ways (such as carry, hold, ride, park, push \etc), but the relationships between bicycles and other objects are far simpler.  The human interactions also imply intent, and possibly provide information about the past or future that is typically lacking from object-object realtionships.
Previous work \cite{chao2017learning, chao2015hico} has similarly recognised that human-object interactions 
of particular interest, and have proposed several datasets. %

As in so many problems of practical interest, the label space of realistic HCVRD exhibits a long tail distribution, meaning that there are very few, to zero, training examples for the vast majority of labels.  This is a fundamental problem for the standard deep learning approach, which relies on large numbers of examples for each class.  If deep learning is to progress from easy, and often artificially simplified problems for which copious training data is available, datasets will need to better reflect the practical reality of the majority of problems.
The main contribution of this paper is a large-scale human-centric visual relationships detection dataset (HCVRD), which accurately depicts the long-tail label distribution of the problem, thus necessitating zero-shot recognition. 

We formulate the human-centric visual relationships detection problem as that of detecting  relationship triplets $\left\langle \emph{human, predicate, object} \right\rangle$ in the image, with bounding boxes on the human subject and object. The HCVRD dataset is constructed based on the Visual Genome \cite{krishna2016visual}. Compared to the previous \textit{human-object interaction} works \cite{chao2017learning, chao2015hico}, there are several differences. First, we have more fine-grained labels. For the `human' item in the triplet, we are not satisfied only detecting a `human' subject, instead, we have four sub-categories which are man(adult), woman(adult), boy and girl. This is a valuable because the gender and age can affect the way that a humans interact with objects. For example, we are unlikely find `a man holding a Barbie' and but this relationship is more commonly seen for `a girl'. Except for the `human' type, our `predicate' covers a much wider range of `relationships' than the `interactions' in the previous setting. %
The dataset contains 9852 different relationships, nearly 20 times morethan the HICO dataset \cite{chao2015hico}. Such big label space leads to a long-tail label distribution, \ie some labels appear less than 10 times. Additionally, we provide 18,471 zero-shot relationships, \ie relationships that never appear in the training split. To the best of our knowledge, this is the biggest dataset with these two forms of label provided and that is labeled with both human-centric visual relationships and corresponding `human' and `object' bounding boxes.

Motivated by the above challenges, our second contribution is developing methods for (i) automatically augmenting the training set using weakly labelled data trawled from the web; and (ii) performing zero-shot recognition by comparing the query data to web-retrieved data.  While not radically novel in approach, our methods address the issues raised in long-tail datasets and provide, we believe, a strong baseline for further work based on our HVCRD dataset and similar data.

%% file: relwork.tex
\vspace{-8pt}
\section{Related work}
\vspace{-8pt}
Visual relationship detection \cite{li2017vip, Liang2017VRD,lu2016visual, zhang2017visual, zhuang2017towards} has attracted a lot of attention, thanks to the fast development of the object detection~\cite{deng2009imagenet}, action recognition~\cite{yao2011human} and so on. Recently, Lu \etal \cite{lu2016visual} proposed a model that uses language priors from semantic word embeddings to finetune the likelihood of a predicted relationship. %
The recently released Visual Genome dataset \cite{krishna2016visual} provides a large-scale annotation of images containing large numbers of objects, attributes and relationships. All these works are interested in the visual relationships between arbitrary two objects in the image. Although this direction is quite interesting and challenging, our focus is different. We are more care about human object, \ie we aims to detect human-centric visual relationships, which has more significant practical and research demands. In practical, human-centric photos account for a large portion of images on the Internet. At the research side, human-object relationships cover a wider range of categories including verbs, actions, spatial and so on.%

Our work is highly related to the studies of human-object interactions (HOI)~\cite{chao2015hico}, which mainly focus on learning the human actions on an object. Earlier methods, such as \cite{gupta2009observing, yao2010modeling, yao2012recognizing} develop joint models of body pose configuration and object location within the image. Yao and Feifei \cite{yao2010grouplet} learns spatial groupings of low-level (SIFT) features for recognizing human-object interactions. Delaitre \etal \cite{delaitre2011learning} introduce a person-object interaction feature representation based on spatial co-occurrences of individual body parts and objects while \cite{desai2010discriminative, hu2013recognising} learn a discriminative model. Prest \etal \cite{prest2012weakly} propose a model that is learned from a set of images annotated only with the action label.

Chao \etal \cite{chao2015hico} introduce a Humans Interacting with Common Objects (HICO) dataset which contains 117 actions and 80 objects. Each image in the dataset contains only one label of $\left\langle \emph{action, object} \right\rangle$ and there are totally 520 such pair categories. Most recently, the HICO-DET \cite{chao2017learning}, an incremental version of the HICO dataset is proposed. In the HICO-DET, the bounding boxes of the human and the objects in an image are annotated, so that the dataset is suitable for a detection task. Gupta \etal \cite{gupta2015visual} provide a Verbs-COCO dataset which has similar settings with HICO-DET. There are 26 action categories and 80 object classes. They allow there are multiple persons in a single image, but they restrict that one person can only has one type of action on one object. Our HCVRD dataset has no such restrictions, thus, we have at least one human in the image, and each human can have multiple relationships with multiple objects. Moreover, we do not restrict that the relationship between the human and an object must be an verb or action, we provide a rich set of predicates, which has more than 900 categories. Consider we have 1,824 object categories, we finally have more than 9000 relationship triplets $\left\langle \emph{human, predicate, object} \right\rangle$. A such big label space leads to a long-tail distribution problem of the data, \ie some classes may have thousands of training examples while some only have very few (less than 10). In this case, the previous fully supervised model \cite{gupta2009observing, yao2010grouplet, chao2017learning} will have issues to learn an effective model. In this paper, we provide a webly-supervised baseline model that learns from large numbers of web returned noisy data, in order to address the long-tail issues and the zero-shot problems. We perform especially well when the training data is lacking.

Our work is also related to the few-shot and zero-shot learning. The few-shot learning \cite{ravi2017optimization} problem focus on when training sets only contain few labeled examples while the zero-shot learning \cite{lampert2014attribute} aims to recognise objects whose instances may not have been seen during training. A more comprehensive review about the zero-shot learning can be found in \cite{xian2017zero}. The learning from imbalanced data \cite{he2009learning} is also related to our work.

%% file: data.tex
\section{The HCVRD dataset}
\label{sec:dataset}

\subsection{Constructing HCVRD dataset}
Our proposed human-centric visual relationship detection (HCVRD) dataset is constructed based on the Visual Genome dataset \cite{krishna2016visual}, which provides detailed scene annotations, such as objects, attributes and relationships. Since we are only interested in the relationships occurred on the human subject, the first step is to extract all the human-related relationships from the 2.3 million relationships pool in the Visual Genome \cite{krishna2016visual}. It is worth noting that there are some relationships only appear once in the dataset. We annotate an `zero-shot' tag on those labels so that they can test under the zero-shot setting. This is one of the significant differences with previous human-object interaction dataset, such as the HICO~\cite{chao2017learning}. The zero-shot setting can verify the generalization ability of an algorithm, \ie the ability to detect unseen relationships in the training set.

The collected relationships are still noisy and should be carefully processed. We first manually correct the annotations that contain misspellings and noisy characters (e.g. comma). We then eliminate the attribute predicates (such as ``has'', ``is'', ``are'') because these predicates are too abstract and may lead to a weak discriminative model. We further normalize the predicates and finally have 927 categories, which cover a wide range of types, such as action, spatial, preposition, comparative and verb and so on. We then merge some semantically similar objects by using the GloVe~\cite{pennington2014glove} and normalize the remaining object names while keeping their fine-grained attributes (e.g. black shirts, yellow shirts). %
Furthermore, we divide the `human' subject into four more fine-grained classes, which are man(adult), woman(adult), boy and girl. This is a valuable setting because the gender and age can affect the way that the human interact with objects.

\vspace{-2mm}
\subsection{Dataset Statistics}
\vspace{-3mm}
Table \ref{tab:re-rank} provides summary statistics about our proposed HCVRD dataset, compared with some human-object interactions dataset. In the following part, we highlight several interesting aspects of the data.

\begin{table}[htp]
	\centering
  \resizebox{1.00998\linewidth}{!}
	{
		\begin{tabular}{c |c |c |c |c | c}
			\hline
		    Datasets &\#relationships (no zero-shot) &\#predicates &\#objects  &\#images &\#zero-shot relationships\\\hline
		    Verbs-COCO~\cite{gupta2015visual}  &- &26 &80 &10346 &-\\\hline
		    Stanford 40 actions~\cite{yao2011human} &40 &35 &28 &9532 &- \\\hline
		    MPII Human Pose~\cite{andriluka20142d} &410 &- &66 &40522 &- \\\hline
			HICO-DET~\cite{chao2017learning} &520  &117  &80  &47774 &-\\\hline
			Ours &9852 &927  &1824  &52855  &18471\\\hline
		\end{tabular}
	}
  \vspace{.2cm}
	\caption{Comparison of the existing human-object interaction detection datasets.}
	\label{tab:re-rank}
\end{table}

\begin{figure}
\centering
\begin{minipage}{.2\textwidth}
	\centering
	\caption{The long-tail label distribution of the dataset. We use the top-2000 relationship types in this figure.}
	\label{fig:distribution}
\end{minipage} %
\begin{minipage}{.75\textwidth}
	\centering
	\includegraphics[width=100mm, height=40mm]{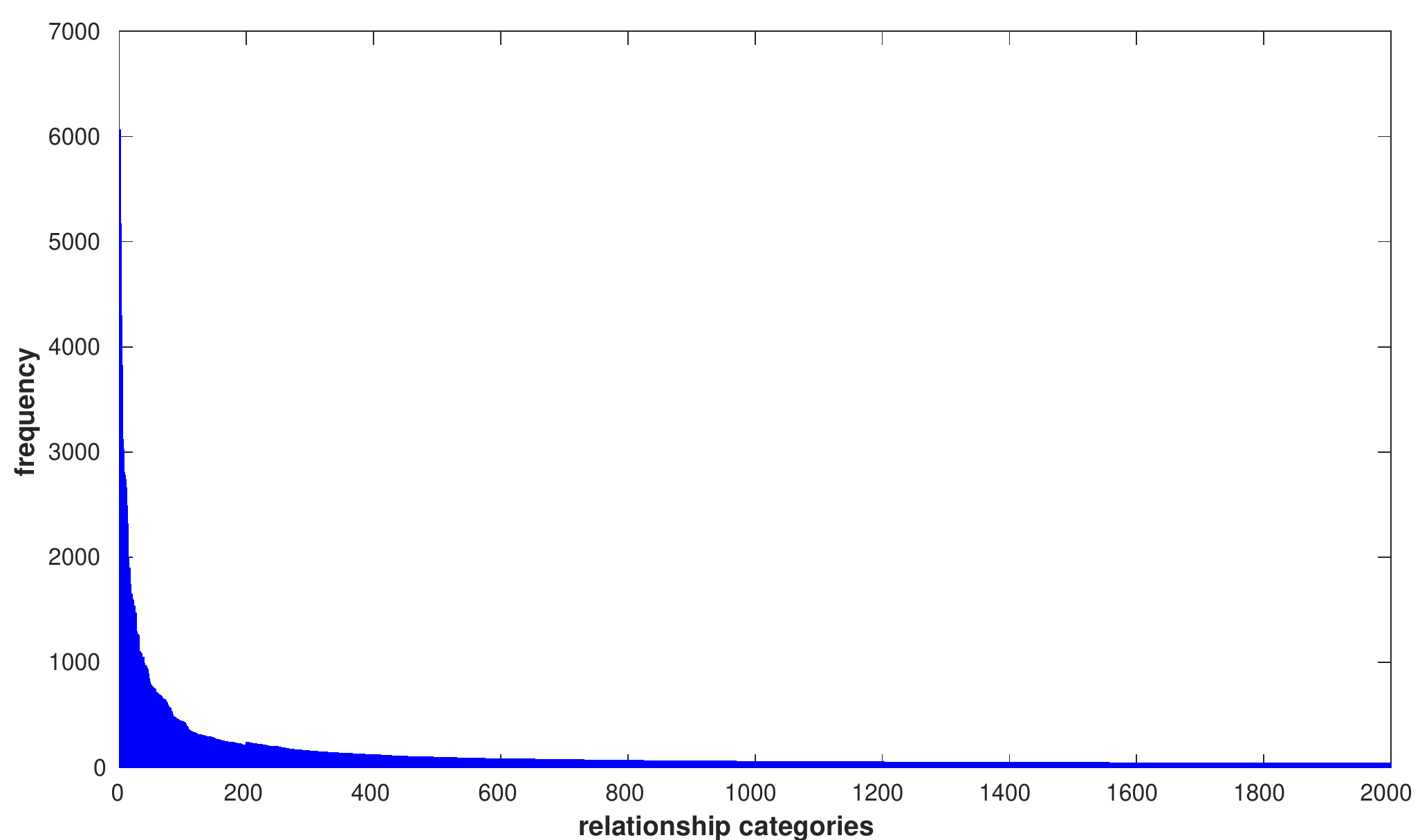}
\end{minipage}
\end{figure}

We finally have 52,855 images with 1,824 object categories and 927 predicates. In total, the dataset contains 256,550 relationships instances with 9,852 non zero-shot relationship types and 18,471 zero-shot relationships types. There are 10.63 predicates per object categories.
We use 31,586 images for training and construct two test splits. The first test split contains 10,000 images where all the relationships occur in the training set. Another test split includes all the zero-shot relationships. %
The distribution of human-object relationships in our dataset (see Figure~\ref{fig:distribution}) highlight the long-tail effect of infrequent relationships. Specifically, there are 370 relationships that appear more than 100 times and 7,474 relationships appear less than 10 times.

\begin{figure*}[tbp]
	\centering
	\resizebox{0.9\linewidth}{!}
	{\begin{tabular}{c}
			\includegraphics{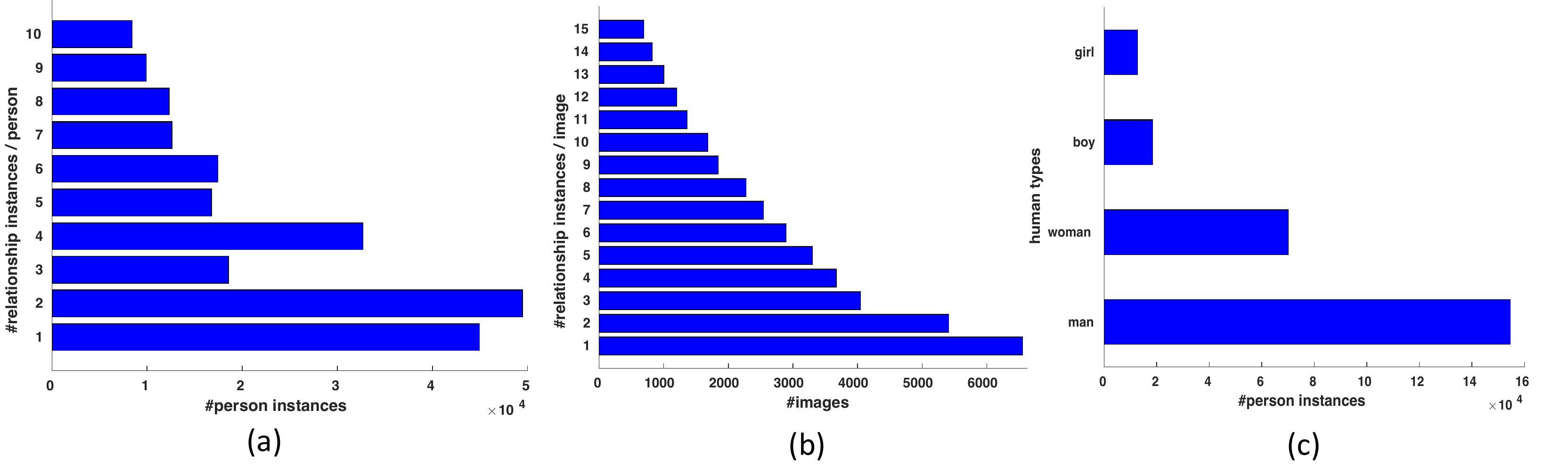}
		\end{tabular}
	}
	\caption{Statistics of the HCVRD dataset, the distribution of the (a): number of relationships in each image. (b): number of different relationships that occurred on a person. (c): human types.}
	\label{fig:statistics}
	\vspace{-10pt}
\end{figure*}

Figure~\ref{fig:statistics} (a) shows the distribution of number of relationship instances in each image. Our HCVRD dataset has a large number of images with more than one relationship instance. On average there are 6.13 relationship instances annotated per image. Figure~\ref{fig:statistics} (b) shows a distribution of the number of different relationship instances that occurred on a person. Unlike past datasets where each person only can have one relationship, each people in the HCVRD dataset has on average 2.62 relationships with other objects. Figure~\ref{fig:statistics} (c) shows the distribution of human types (such as man, woman, boy and girl) in our dataset.

%% file: method.tex
\vspace{-3mm}
\section{A webly-supervised model}
\label{sec:method}

One of the biggest challenges in our proposed dataset is the long-tail distribution of the labels, as shown in the Figure \ref{fig:distribution}. Nearly 80\% of the relationship labels in our dataset have less than 10 training examples. This issue casts a big challenge to the conventional 
supervised learning models, especially for those deep convolutional neural network based models, which normally require a large number of examples to train. In this paper, we propose a webly-supervised learning approach to effectively utilize the unlimited web data for recognizing human-centric relationships. Our model is not restricted by the limited training examples and the long-tail distribution issue.

\begin{figure}[tbp]
	\centering
	\resizebox{0.8\linewidth}{!}
	{\includegraphics{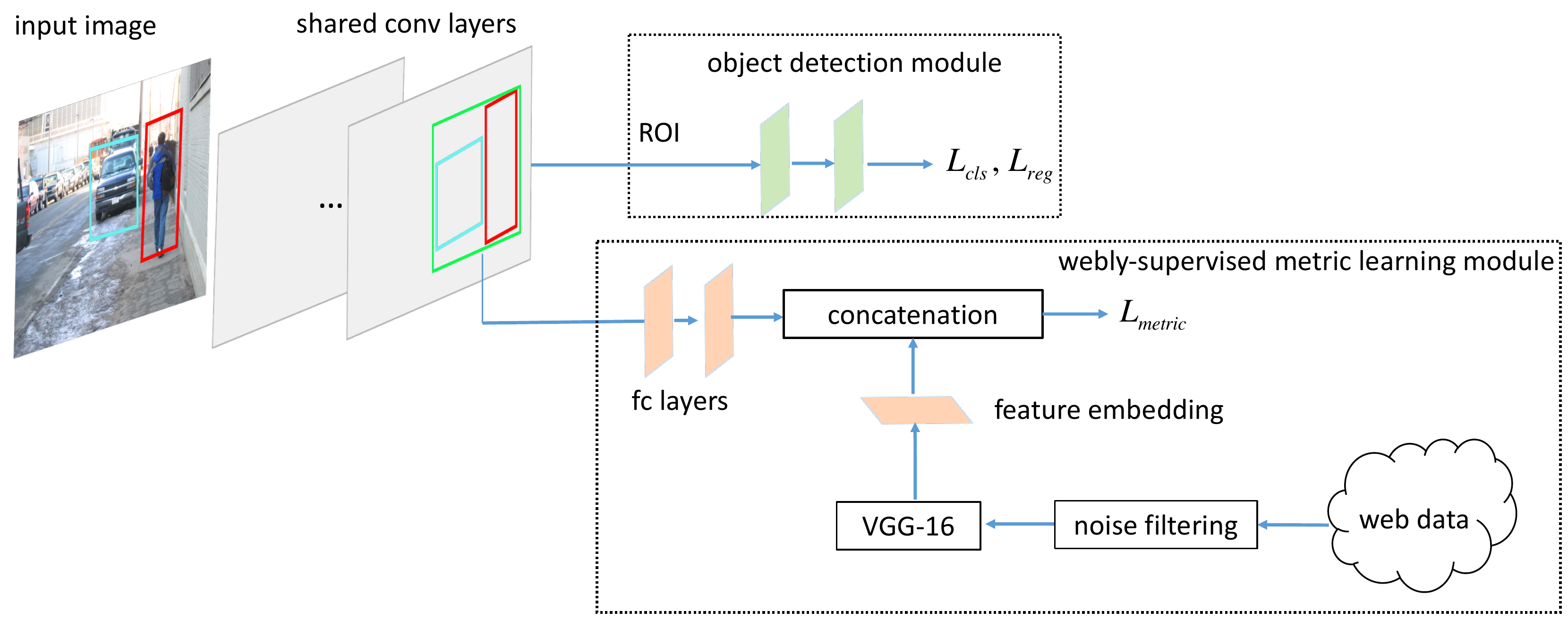}}
	\caption{The framework of the proposed model. The model consists of (a): an object detection module, (b) a webly-supervised metric learning module. The two modules can be jointed trained end-to-end.}
   \label{fig:overview}
   \vspace{-15pt}
\end{figure}

An overview of our proposed webly-supervised relationship detection (WSRD) model is shown in Figure~\ref{fig:overview}, our model is divided into two submodules: the detection module and the distance metric learning module. The detection module is in the style of Faster-RCNN \cite{ren2015faster}, which is used to detect the object and human subject (in subcategory). The union regions of the detected human-object pair are sent to a deep metric learning module, which computes the distance between the proposed region and the web crawled visual relationships data, in order to decide the `predicate' category between the detected `human' and the `object'. The two submodules can be learned in an end-to-end manner and share the convolutional layers for efficiency. The proposed model takes the entire image as input and fed it into several convolutional and max-pooling layers to get the feature map. The VGG-16~\cite{simonyan2014very} network is used as basic building blocks for our model. We discuss the two submodules in the following part. 
	
\subsection{Detection module} 
The object (and human subject) detection module structure is identical to that of the Faster-RCNN~\cite{ren2015faster}. Taking the output of the \texttt{Conv5\_3} feature map as the input, the Region Proposal Network (RPN) is used to generate object proposals. During training, we extract features with RoIPool for each object proposal, followed by the bounding box regression loss ${L_{reg}}$ and a classification loss ${L_{cls}}$ to learn the detector which are identical as those defined in \cite{ren2015faster}. During the inference, we first detect all human subjects and objects in the images. To remove the redundancy, we apply the non-maximum suppression (NMS) method to reduce the number of proposals with the IoU threshold 0.3 and objectiveness scores higher than 0.2. These filtered boxes are further grouped to $\left\langle \emph{human, object} \right\rangle$ pairs and are used as the input to the distance metric learning module.

\subsection{Distance metric learning module}

In this module, we accept the union regions of the detected human and object and compute the distance between the proposed union region and the web crawled visual relationships data. The nearest class label of the web data will be assigned to the proposed region. The distance measure function is learned by applying a deep metric learning module, which can be further divided into following 3 steps:
	
\emph{Web data crawling}:
To collect images for the metric learning, we automatically crawl from Google Images as the source of candidate images. For most basic categories commonly appeared in the vision world, the top results returned by Google image search are quite clean so that we can learn useful visual representations from them. We treat all the 9,852 relationships as the query list and process each of the category independently. 100 images are crawled from the Google Images for each single relationship class.

\emph{Weakly-supervised filtering}
However, the returned images of some relationship categories may contain massive noise and will adversely affect the model. To mitigate this issue, we employ a weakly-supervised noise robust approach \cite{zhuang2016attend} to filter the noisy images. The noise filtering process relies on a random group training process that randomly group multiple web data into a single training instance as the input of an classification neural network. Then an attentional pooling layer is employed on the last convoultional layer to select parts of the neuron activations and pools the activations into the instance-level representation for classification. Once the network is trained, the random grouping module can be dropped off and the attention weights decide the confidence score of each training example. We then sort the confidence scores in descending order for each category. In terms of this case, the high-ranking images are correctly labeled ones and vice versa. We empirically keep the top 80 \%  images as the web training data and discard the remaining data. 
	
\emph{Deep metric learning}
We intend to learn the semantic feature embeddings where similar examples are mapped close to each other while dissimilar examples are mapped farther apart. By learning such distance metric, we can map a human-object region to its nearest relationship category in a webly-supervised manner. After obtaining the web data, we first construct a set of positive pairs and a set of negative pairs incrementally in the following manner: when a pair $({{\bf{x}}_i}, {{\bf{x}}_j})$ is from the same class, we treat it as a positive pair and vice versa. In our approach, sample ${{\bf{x}}_i}$ is from the HCVRD dataset and ${{\bf{x}}_j}$ is from the web data. 

We then denote the triplet relationship by $\bf{t} = \left\langle {{\bf{b}}_s, {\bf{b}}_p, {\bf{b}}_o} \right\rangle$, where ${\bf{b}}_s$, ${\bf{b}}_p$ and ${\bf{b}}_o$ are the bounding boxes for subject, predicate and object, respectively. The predicate region ${\bf{b}}_p$ is the union of ${\bf{b}}_s$ and ${\bf{b}}_o$. During the training, the ground truth predicate region in \texttt{Conv5\_3} is used as part of the input for the metric learning module. In the inference, we first detect the human and objects and get all the possible union bounding boxes. And then the signal is returned to the \texttt{Conv5\_3} layer (the last convolutional layer) of the VGG-16 to obtain the corresponding union region feature for ${\bf{b}}_p$, which is sent to two fully connected layers and the output serves as part of the input for the metric learning module. Another input is from the collected web data, which is passed through a pre-trained VGG-16 model and a learnable feature embedding layer. Two embedded features are then concatenated together as the final pairwise representation. 
Following \cite{oh2016deep}, the metric is learned using a structured loss function based on the sampled positive and negative pairs of training samples:
\begin{equation}
\begin{array}{l}
L_{mec} = \frac{1}{{2\left| \P \right|}}\sum\limits_{(i,j) \in \P} {\max (0,\,\,{L_{i,j}}} {)^2},\\
{L_{i,j}} = \log (\sum\limits_{(i,k) \in \N} {\exp (\alpha  - {D_{i,k}})} ,\,\sum\limits_{(j,k) \in \N} {\exp (\alpha  - {D_{j,l}})} ) + {D_{i,j}}
\end{array}	
\end{equation}
where $\P$ is the set of positive pairs and $\N$ is the set of negative pairs, ${D_{i,j}} = {\left\| {f({x_i}) - f({x_j})} \right\|_2}$ is distance between two embedding feature vectors. 

The two modules can be jointly trained in an end-to-end manner.  The model employs multi-task loss for human-object relationship detection:
\vspace{-1mm}
\begin{equation}
L = {L_{reg}} + {L_{cls}} + {L_{mec}}
\end{equation}
where $L_{reg}$ and $L_{cls}$ are the regression loss and cross-entropy loss in the Faster-RCNN module.

%% file: exp.tex
\section{Experiments}
\label{exp}
\subsection{Implementation details}

We set the feature embedding size in the metric learning module as 256. For training efficiency, we first pretrained the detection module and fix it while training the metric learning module. The learning rate is initialized to 0.0001 and decreased by a factor of 10 after every 5 epochs. During the inference, we first retrieve the top 20 nearest neighbor relationships and select those including both detected human and object categories. Then we use the top-ranked selected candidates for evaluation.

\subsection{Evaluation setup}
We evaluate our human-object interactions task using Recall@100 and Recall@50, following the setting of Visual Relationship Detection (VRD) task~\cite{Liang2017VRD, lu2016visual}. Recall@x computes the fraction of times the correct relationship is calculated in the top x predictions, which are ranked by the final distances. We evaluate on three tasks: (1) For \textbf{predicate detection}, the goal is to predict the accuracy of \emph{predicate} recognition, where the labels and bounding boxes for both the \emph{object} and \emph{human} are given. (2) In \textbf{phrase detection}, we aim to predict $\left\langle \emph{human-predicate-object} \right\rangle$ and localize the entire relationship in one bounding box. (3) For \textbf{relationship detection}, the task is to recognize $\left\langle \emph{human-predicate-object} \right\rangle$
and localize both human and object bounding boxes, where both boxes should have at least 0.5 overlap with the ground-truth in order to be regarded as correct prediction.
10
In the real world applications, different relationships may share very similar semantic meanings (e.g. ``man holding phone'', ``man talking on phone'',  ``man using phone'') and it's difficult to differentiate them. Hence, in many cases, one ``appropriate'' prediction may be judged ``incorrect'' due to the limitation of the test annotations, which is a common problem of the current VRD evaluation metric. One possible solution is to employ the human evaluation, which is cost however. In this paper, we instead report both top-1 and top-3 results under different Recalls to evaluate the model. %
\subsection{Baselines}
We benchmark the following approaches on our new dataset.

\textbf{\emph{Multilabel classification}}  A person can concurrently perform different interactions with different target objects, e.g. a person can ``ride bicycle'' and ``drink water''  at the same time.  Thus we treat the human-object relationship detection task as a multilabel classification problem where we apply a sigmoid cross entropy loss on top of the classification layer. Specifically, we treat the union of a human and its correlated objects as the input during training. During the testing, we use our object detection module to return the regions. We use VGG-16 model as the basis building block.

\textbf{\emph{JointCNN}}
This implements the Visual phrases~\cite{sadeghi2011recognition}. We train a VGG-16 model to jointly predict the three components of a relationship. Specifically, we treat each relationship category separately and train a 9,852 way classification model.

\textbf{\emph{SeparateCNN}}
Following the visual model of \cite{lu2016visual}, we first train a VGG16 model to classify the 1,824 objects. Similarly, we train a second model to classify each of the 927 predicates using the union of the bounding boxes of the participating human and the object in that relationship.

\begin{table}[tbp]
	\centering
	\scalebox{0.875}
	{
		\begin{tabular}{c|c c| c c| c c| c c| c c| c c}
			\hline
			\multirow{3}{*}{Method}&\multicolumn{4}{c|}{Predicate Det.} &\multicolumn{4}{c|}{Phrase Det.}
			&\multicolumn{4}{c}{Relationship Det.} \\  &\multicolumn{2}{c|}{R@50}&\multicolumn{2}{c|}{R@100}&\multicolumn{2}{c|}{R@50}&\multicolumn{2}{c|}{R@100}&\multicolumn{2}{c|}{R@50}&\multicolumn{2}{c}{R@100} \\
			&top-1 &top-3 &top-1 &top-3 &top-1 &top-3 &top-1 &top-3 &top-1 & top-3 &top-1 & top-3\\\hline
			Multilabel &0.87  &2.78  &0.87  &2.78 &0.44 &0.92 &0.50 &0.95 &0.03  &0.07  &0.04  &0.09 \\
			JointCNN &2.68  &7.36 &2.68  &7.36  &2.35 &5.63  &2.39 &6.14 &0.21  &0.44 &0.22  &0.53 \\
		    SeparateCNN &29.00  &44.37  &29.00  &45.87 &8.24  &10.53  &8.92  &13.81  &0.48  &0.60  &0.50  &0.66 \\
			Ours &31.08  &47.66 &31.08  &48.98 &10.03  &13.05  &10.75 &16.94 &0.53  &0.68  &0.59  &0.72 \\\hline

		\end{tabular}
  }
  \vspace{.2cm}
		\caption{Evaluation of different methods on the proposed dataset. The results reported include visual relationship detection (Relationship Det.) and predicate detection (Predicate Det.) measured by Top-100 recall (R@100) and Top-50 recall (R@50).}
		\label{tab:relationship}
	\end{table}

For \emph{JointCNN} and \emph{Multilabel} baselines, we empirically find that due to the long-tail property of the dataset, the learned models are seriously biased. It causes the predictions only fall into those labels that with large numbers of training examples. To solve the problem of extreme classification with enormous number of categories, we instead propose to employ the metric learning approach with web data to perform efficient nearest neighbor inference on the learned metric space. By comparing \emph{ours} with the two baselines, we find significant performance increase on all evaluation metrics.
For the \emph{SeperateCNN} baseline, since the training data for human, objects and predicates are relatively adequate respectively, its performance is competitive with our proposed method.
However, compared to predicate detection results, the performance of phrase and relationship detection decreases a lot. It shows that detecting such wide range of objects is a major challenge for visual relationship detection.
We also show some qualitative examples in Figure~\ref{fig:qualitative} and more examples are provided in the supplementary material.

\subsection{Long-tail evaluation}
Due to the long-tail distribution of the categories in the dataset, the infrequent relationships will contribute not much to the final testing performance. But in real world applications, the relationships in long-tail should not be ignored. So we select those relationships that appear less than 10 times as a subset (i.e. there are totally 7,474 relationships) and report the performance in Table~\ref{tab:longtail}. From the table, we can see that our approach performs steadily better than the baseline methods. For the baseline methods, the lack of training data is a main challenge for obtaining accurate predictions. The main motivation of the proposed method is to utilize web data to tackle this limitation. With the always available web data, we can learn the general concept of distance metrics and efficiently infer nearest neighbor relationships on the learned metric space.

\begin{table}[H]
	\centering
	\scalebox{0.7}
	{
		\begin{tabular}{c|c c| c c| c c| c c|c c |cc}
			\hline
			\multirow{3}{*}{Method}&\multicolumn{4}{c|}{Predicate Det.} &\multicolumn{4}{c|}{Phrase Det.}
			&\multicolumn{4}{c}{Relationship Det.}  \\  &\multicolumn{2}{c|}{R@50}&\multicolumn{2}{c|}{R@100}&\multicolumn{2}{c|}{R@50}&\multicolumn{2}{c|}{R@100}&\multicolumn{2}{c|}{R@50}&\multicolumn{2}{c}{R@100} \\
			&top-1 &top-3 &top-1 &top-3 &top-1 &top-3 &top-1 &top-3 &top-1 &top-3 &top-1 &top-3\\\hline
			Multilabel &0.45  &1.09  &0.45  &1.09 &0.22 &0.58 &0.24 &0.62   &0.01  &0.01  &0.01   &0.01  \\
			JointCNN &0.02  &0.03  &0.02  &0.03 &0.01 &0.01 &0.01 &0.01 &0.01   &0.01  &0.01  &0.01   \\
			SeparateCNN &15.94   &26.73  &15.94 &26.73 &0.49 &1.55 &0.58  &1.96  &0.04  &0.08  &0.05  &0.10  \\
			Ours-without web data &18.01 &29.35 &18.01 &29.35 &0.73 &2.15 &0.80 &2.43 &0.06 &0.10 &0.07 &0.13\\
			Ours &24.55  &36.59   &24.55  &36.59  &1.76 &3.62 &1.91 &4.56  &0.12   &0.16   &0.14   &0.21   \\\hline

		\end{tabular}}
		\caption{Results for human-object relationship detection on the long-tail benchmark subset.}
		\label{tab:longtail}
	\end{table}

\subsection{Ablation study}
In this section,  we evaluate the contributions of different factors in our system to the results.

\emph{With vs. without metric learning module}
Metric learning module is the key component of our system. To evaluate its impact, we implement a variant without the metric learning module. For the detected union bounding boxes of relationships and web data, we directly extract the 4096-dimentional feature vector for each sample using the pretrained VGG-16 model. We then compute the cosine similarity between the test sample and all mean vectors of the relationship categories that contain both detected human and object types. We then retrieve the nearest neighbor relationship categories as our predictions.
Table~\ref{tab:ablation} (a) vs. (c) shows that learning the semantic feature embeddings via distance metric contributes a lot to the final performance.

\emph{With vs.\ without web data}
We also evaluate the influence of the web data by only using the training data of the dataset. Since one motivation of introducing web data is to solve the scarceness of training data, we report this variant under the long-tail setting in Table~\ref{tab:longtail} as \emph{Ours-without web data}. By comparing it with \emph{Ours} in Table~\ref{tab:longtail},  we find that removing web data causes an obvious performance degradation, which proves the effectiveness of introducing the web data.

\emph{With vs. without noise filtering}
We further remove the noise filtering step to investigate the affect of noisy labels. The results are shown in Table~\ref{tab:ablation} (b).
Table~\ref{tab:ablation} (b) vs. (c) shows that removing noise filtering have less affect to the performance compared to removing metric learning module. This is because for relationships that commonly used in the vision community, top results returned by Google images search are pretty clean. Noise filtering provides an auxiliary to further improves the quality of web data.

\vspace{-3mm}
\begin{table}[H]
	\centering
	\scalebox{0.7}
	{
		\begin{tabular}{c|c c| c c| c c| c c | c c | c c}
			\hline
			\multirow{3}{*}{Method}&\multicolumn{4}{c|}{Predicate Det.} &\multicolumn{4}{c|}{Phrase Det.}
			&\multicolumn{4}{c}{Relationship Det.}  \\  &\multicolumn{2}{c|}{R@50}&\multicolumn{2}{c|}{R@100}&\multicolumn{2}{c|}{R@50}&\multicolumn{2}{c|}{R@100}&\multicolumn{2}{c|}{R@50}&\multicolumn{2}{c}{R@100} \\
			&top-1 &top-3 &top-1 &top-3 &top-1 &top-3 &top-1 &top-3 &top-1 &top-3 &top-1 &top-3\\\hline
			(a) Without metric learning module &22.55  &33.12  &22.55  &33.87 &5.87 &7.33 &6.04 &9.44 &0.29  &0.43   &0.34   &0.49  \\
			(b) Without noise filtering &30.36  &46.12  &30.37  &46.68 &9.92 &12.96 &10.67 &16.36  &0.49  &0.64  &0.57  &0.70  \\
			(c) Ours (full model) &31.08  &47.66  &31.08  &48.98 &10.03 &13.05 &10.75 &16.94 &0.53  &0.68  &0.59  &0.72  \\\hline

		\end{tabular}}
		\caption{Ablation studies on the HCVRD benchmark non-zeroshot test set.}
		\label{tab:ablation}
	\end{table}
\subsection{Zero-shot evaluation}
It is quite important to make the model generalizable to unseen human-object relationships. In this section, we report the performance of our method on a zero-shot learning setting. Specifically, we train our models on the training set and evaluate their relationship detection performance on the $18,471$ unseen visual relationships in the zero-shot test split. Given the detected human and objects in a relationship, we first get all their possible interactions to get a search space. We then collect web data and extract feature embeddings to get the nearest neighbors relationships for the test sample. The results are reported in Table~\ref{tab:zeroshot}. From the table, we can see that the proposed method works more robust. This can be attributed to the introduction of the external web data for efficient nearest neighbor search. However, one main limitation of our method is the huge computational complexity on the zero-shot test setting. For the ``separateCNN'' baseline, by predicting the interactions seperately from its objects, it is difficult to capture the appearance variations due to the weak and even ambiguous visual features.

\begin{table}[H]
	\centering
	\scalebox{0.75}
	{
		\begin{tabular}{c|c c| c c| c c| c c|c c|cc}
			\hline
			\multirow{3}{*}{Method}&\multicolumn{4}{c|}{Predicate Det.} &\multicolumn{4}{c|}{Phrase Det.} &\multicolumn{4}{c}{Relationship Det.}  \\  &\multicolumn{2}{c|}{R@50}&\multicolumn{2}{c|}{R@100}&\multicolumn{2}{c|}{R@50}&\multicolumn{2}{c|}{R@100} &\multicolumn{2}{c|}{R@50}&\multicolumn{2}{c}{R@100} \\
			&top-1 &top-3 &top-1 &top-3 &top-1 &top-3 &top-1 &top-3 &top-1 &top-3 &top-1 &top-3\\\hline
			Multilabel &-   &-  &-   &-   &-   &-   &-   &-  &- &- &- &-\\
			JointCNN &-  &-  &-  &-  &-   &-  &-  &-  &- &- &- &-\\
			SeparateCNN &2.75  &4.98  &2.99  &5.93 &0.06  &0.11 &0.07 &0.16  &0.01  &0.05  &0.03  &0.08 \\
			Ours &8.15  &12.34  &8.57  &13.42  &0.88 &1.43 &0.92 &1.84  &0.03  &0.09  &0.05  &0.12  \\\hline

		\end{tabular}}
		\caption{Results for human-object relationship detection on the zero-shot benchmark test set.}
		\label{tab:zeroshot}
	\end{table}

\begin{figure}[tbp]
	\centering
	\resizebox{0.99\linewidth}{!}
	{\includegraphics{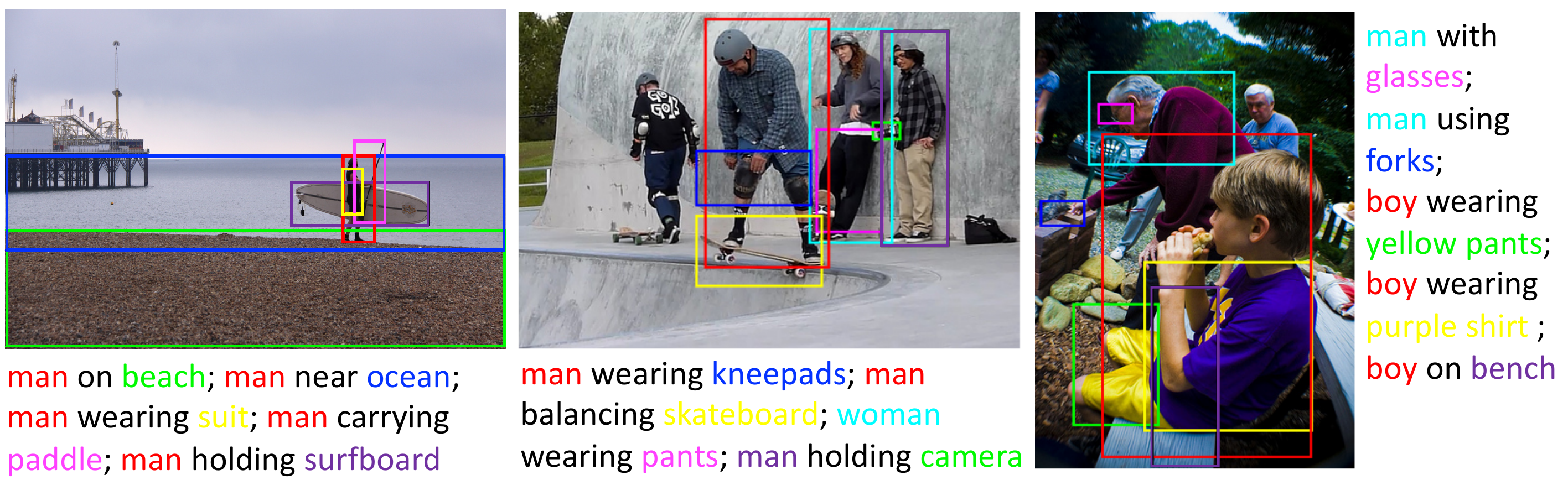}}
	\caption{Qualitative examples of the predicate detection. The color of human and objects in the phrases correspond to the color of the bounding boxes. We only predict the interactions between the ground-truth bounding box pairs.}
	\label{fig:qualitative}
\end{figure}

%% file: nips_2017.bbl
\begin{thebibliography}{10}

\bibitem{andriluka20142d}
M.~Andriluka, L.~Pishchulin, P.~Gehler, and B.~Schiele.
\newblock 2d human pose estimation: New benchmark and state of the art
  analysis.
\newblock In {\em Proc. IEEE Conf. Comp. Vis. Patt. Recogn.}, 2014.

\bibitem{chao2017learning}
Y.-W. Chao, Y.~Liu, X.~Liu, H.~Zeng, and J.~Deng.
\newblock Learning to detect human-object interactions.
\newblock {\em arXiv preprint arXiv:1702.05448}, 2017.

\bibitem{chao2015hico}
Y.-W. Chao, Z.~Wang, Y.~He, J.~Wang, and J.~Deng.
\newblock Hico: A benchmark for recognizing human-object interactions in
  images.
\newblock In {\em Proc. IEEE Int. Conf. Comp. Vis.}, 2015.

\bibitem{delaitre2011learning}
V.~Delaitre, J.~Sivic, and I.~Laptev.
\newblock Learning person-object interactions for action recognition in still
  images.
\newblock In {\em Proc. Adv. Neural Inf. Process. Syst.}, 2011.

\bibitem{deng2009imagenet}
J.~Deng, W.~Dong, R.~Socher, L.-J. Li, K.~Li, and L.~Fei-Fei.
\newblock Imagenet: A large-scale hierarchical image database.
\newblock In {\em Proc. IEEE Conf. Comp. Vis. Patt. Recogn.}, 2009.

\bibitem{desai2010discriminative}
C.~Desai, D.~Ramanan, and C.~Fowlkes.
\newblock Discriminative models for static human-object interactions.
\newblock In {\em CVPR workshops}, 2010.

\bibitem{gupta2009observing}
A.~Gupta, A.~Kembhavi, and L.~S. Davis.
\newblock Observing human-object interactions: Using spatial and functional
  compatibility for recognition.
\newblock {\em IEEE Transactions on Pattern Analysis and Machine Intelligence},
  31(10):1775--1789, 2009.

\bibitem{gupta2015visual}
S.~Gupta and J.~Malik.
\newblock Visual semantic role labeling.
\newblock {\em arXiv preprint arXiv:1505.04474}, 2015.

\bibitem{he2009learning}
H.~He and E.~A. Garcia.
\newblock Learning from imbalanced data.
\newblock {\em IEEE Transactions on knowledge and data engineering}, 2009.

\bibitem{hu2013recognising}
J.-F. Hu, W.-S. Zheng, J.~Lai, S.~Gong, and T.~Xiang.
\newblock Recognising human-object interaction via exemplar based modelling.
\newblock In {\em Proc. IEEE Int. Conf. Comp. Vis.}, 2013.

\bibitem{krishna2016visual}
R.~Krishna, Y.~Zhu, O.~Groth, J.~Johnson, K.~Hata, J.~Kravitz, S.~Chen,
  Y.~Kalantidis, L.-J. Li, D.~A. Shamma, et~al.
\newblock Visual genome: Connecting language and vision using crowdsourced
  dense image annotations.
\newblock {\em arXiv preprint arXiv:1602.07332}, 2016.

\bibitem{lampert2014attribute}
C.~H. Lampert, H.~Nickisch, and S.~Harmeling.
\newblock Attribute-based classification for zero-shot visual object
  categorization.
\newblock {\em {IEEE} Trans. Pattern Anal. Mach. Intell.}, 2014.

\bibitem{li2017vip}
Y.~Li, W.~Ouyang, and X.~Wang.
\newblock Vip-cnn: A visual phrase reasoning convolutional neural network for
  visual relationship detection.
\newblock In {\em Proc. IEEE Conf. Comp. Vis. Patt. Recogn.}, 2017.

\bibitem{Liang2017VRD}
X.~Liang, L.~Lee, and E.~P.~Xing.
\newblock Deep variation-structured reinforcement learning for visual
  relationship and attribute detection.
\newblock In {\em Proc. IEEE Conf. Comp. Vis. Patt. Recogn.}, 2017.

\bibitem{lu2016visual}
C.~Lu, R.~Krishna, M.~Bernstein, and L.~Fei-Fei.
\newblock Visual relationship detection with language priors.
\newblock In {\em Proc. Eur. Conf. Comp. Vis.}, 2016.

\bibitem{oh2016deep}
H.~Oh~Song, Y.~Xiang, S.~Jegelka, and S.~Savarese.
\newblock Deep metric learning via lifted structured feature embedding.
\newblock In {\em Proc. IEEE Conf. Comp. Vis. Patt. Recogn.}, 2016.

\bibitem{pennington2014glove}
J.~Pennington, R.~Socher, and C.~D. Manning.
\newblock Glove: Global vectors for word representation.
\newblock In {\em EMNLP}, 2014.

\bibitem{prest2012weakly}
A.~Prest, C.~Schmid, and V.~Ferrari.
\newblock Weakly supervised learning of interactions between humans and
  objects.
\newblock {\em IEEE Transactions on Pattern Analysis and Machine Intelligence},
  34(3):601--614, 2012.

\bibitem{ravi2017optimization}
S.~Ravi and H.~Larochelle.
\newblock Optimization as a model for few-shot learning.
\newblock In {\em Proc. Int. Conf. Learn. Rerp.}, 2017.

\bibitem{ren2015faster}
S.~Ren, K.~He, R.~Girshick, and J.~Sun.
\newblock Faster r-cnn: Towards real-time object detection with region proposal
  networks.
\newblock In {\em Proc. Adv. Neural Inf. Process. Syst.}, 2015.

\bibitem{sadeghi2011recognition}
M.~A. Sadeghi and A.~Farhadi.
\newblock Recognition using visual phrases.
\newblock In {\em Proc. IEEE Conf. Comp. Vis. Patt. Recogn.}, 2011.

\bibitem{simonyan2014very}
K.~Simonyan and A.~Zisserman.
\newblock Very deep convolutional networks for large-scale image recognition.
\newblock In {\em Proc. Int. Conf. Learn. Rerp.}, 2015.

\bibitem{xian2017zero}
Y.~Xian, B.~Schiele, and Z.~Akata.
\newblock Zero-shot learning-the good, the bad and the ugly.
\newblock {\em arXiv preprint arXiv:1703.04394}, 2017.

\bibitem{yao2010grouplet}
B.~Yao and L.~Fei-Fei.
\newblock Grouplet: A structured image representation for recognizing human and
  object interactions.
\newblock In {\em Proc. IEEE Conf. Comp. Vis. Patt. Recogn.}, 2010.

\bibitem{yao2010modeling}
B.~Yao and L.~Fei-Fei.
\newblock Modeling mutual context of object and human pose in human-object
  interaction activities.
\newblock In {\em Proc. IEEE Conf. Comp. Vis. Patt. Recogn.}, 2010.

\bibitem{yao2012recognizing}
B.~Yao and L.~Fei-Fei.
\newblock Recognizing human-object interactions in still images by modeling the
  mutual context of objects and human poses.
\newblock {\em IEEE Transactions on Pattern Analysis and Machine Intelligence},
  34(9):1691--1703, 2012.

\bibitem{yao2011human}
B.~Yao, X.~Jiang, A.~Khosla, A.~L. Lin, L.~Guibas, and L.~Fei-Fei.
\newblock Human action recognition by learning bases of action attributes and
  parts.
\newblock In {\em Proc. IEEE Int. Conf. Comp. Vis.}, 2011.

\bibitem{zhang2017visual}
H.~Zhang, Z.~Kyaw, S.-F. Chang, and T.-S. Chua.
\newblock Visual translation embedding network for visual relation detection.
\newblock In {\em Proc. IEEE Conf. Comp. Vis. Patt. Recogn.}, 2017.

\bibitem{zhuang2016attend}
B.~Zhuang, L.~Liu, Y.~Li, C.~Shen, and I.~Reid.
\newblock Attend in groups: a weakly-supervised deep learning framework for
  learning from web data.
\newblock {\em arXiv preprint arXiv:1611.09960}, 2016.

\bibitem{zhuang2017towards}
B.~Zhuang, L.~Liu, C.~Shen, and I.~Reid.
\newblock Towards context-aware interaction recognition.
\newblock {\em arXiv preprint arXiv:1703.06246}, 2017.

\end{thebibliography}
